\documentclass[11pt]{article}
\usepackage{bigfoot}
\usepackage{multirow}
\usepackage{booktabs}

\usepackage[preprint]{acl}

\usepackage{times}
\usepackage{latexsym}
\usepackage{geometry}
\usepackage[most]{tcolorbox}
\usepackage{amsmath}
\usepackage{amssymb}

\usepackage[T1]{fontenc}
\usepackage{url}
\usepackage{enumitem}

\usepackage[utf8]{inputenc}

\usepackage{microtype}

\usepackage{inconsolata}

\usepackage{graphicx}
\usepackage[table]{xcolor}
\usepackage{cleveref} 
\usepackage{chngcntr}

\newcommand{\deltacol}[1]{%
  \ifdim #1 pt>0pt
    \cellcolor{green!25}\textcolor{green!70!black}{#1}%
  \else
    \cellcolor{red!25}\textcolor{red!70!black}{#1}%
  \fi
}

\title{Do Instruction-Tuned Models Always Perform Better Than Base Models? Evidence from Math and Domain-Shifted Benchmarks}

\author{
 \textbf{Prateek Munjal},
 \textbf{ Clement Christophe},
 \textbf{ Ronnie Rajan},
 \textbf{ Praveenkumar Kanithi}
\\
\\
M42, Abu Dhabi, UAE,
}

\tcbset{
  takeawaybox/.style={
    boxrule=1pt,
    colback=white,
    colframe=black!80!blue,
    arc=3mm,
    boxsep=0mm,
    toptitle=2mm,
    bottomtitle=2mm,
    coltitle=white,
    fonttitle=\bfseries\large,
    halign=left,
    title=Key Takeaway,
    title style={
      fill=black,
      borderline north={1pt}{black!80!blue},
      borderline west={1pt}{black!80!blue},
      borderline east={1pt}{black!80!blue},
      borderline south={0pt}{white}, 
      arc=3mm,
      outer arc=3mm
    },
    left=3mm,
    right=3mm,
    top=2mm,
    bottom=2mm,
    colback=blue!5!white,
  }
}

\newtcolorbox{findingbox}{
    colback=gray!5,       
    colframe=teal!80!black, 
    arc=2mm,              
    boxrule=0.8pt,        
    left=5pt, right=5pt, top=5pt, bottom=5pt, 
    boxsep=1pt
}

\begin{document}
\maketitle
\begin{abstract}

Instruction finetuning is standard practice for improving LLM performance, yet it remains unclear whether it enhances reasoning or merely induces surface-level pattern matching. We investigate this by evaluating base and instruction-tuned models on standard math benchmarks, structurally perturbed variants, and domain-shifted tasks. Our analysis highlights two key (often overlooked) limitations of instruction tuning. First, the performance advantage is unstable and depends heavily on evaluation settings. In zero-shot CoT settings on GSM8K, base models consistently outperform instruction-tuned variants, with drops as high as 32.67\% (Llama3-70B). Instruction-tuned models only match or exceed this performance when provided with few-shot exemplars, suggesting a reliance on specific prompting patterns rather than intrinsic reasoning. Second, tuning gains are brittle under distribution shift. Our results show that base models surpass instruction-tuned variants on the domain-specific MedCalc benchmark. Additionally, instruction-tuned models show sharp declines on perturbed datasets, indicating sensitivity to prompt structure over robust reasoning.

\end{abstract}

\section{Introduction}


Large language models (LLMs) have recently demonstrated impressive performance on a wide range of reasoning benchmarks \citep{grattafiori2024llama, yang2025qwen3, liu2024deepseek, kimi_k2_team2025kimi}. While assessing genuine reasoning capabilities remains challenging, the field primarily relies on solution accuracy across mathematical datasets \citep{gsm8k_cobbe2021training, math_perturb_huang2025math, math500_lightman2023lets} as a proxy. Evaluations on these standard benchmarks consistently suggest that instruction-tuned models significantly outperform their base counterparts \citep{qwen2.5-math-yang2024qwen2}, creating an implicit assumption that instruction tuning universally enhances capability. Consequently, base models are often excluded from comparative analyses or evaluated under restrictive inference settings.


However, this assumption overlooks critical resource trade-offs. While instruction tuning improves format adherence and user interaction, it necessitates massive curated datasets \citep{zhang2024infinitymath, toshniwal2024openmathinstruct, gsm8k_cobbe2021training, liu2024finemath, wei2023cmath, wang2023generative, mitra2024orca, albalak2025big} and substantial training resources \citep{liu2024deepseek}. Simultaneously, test-time strategies such as self-consistency \citep{sefl_consistency_wang2022self}, CoT decoding \citep{cot_decoding_wang2024chain}, ESC \citep{li2024escape} and Pass@k evaluation incur significant computational costs, particularly when sampling multiple trajectories is required to optimize the performance. Despite these costs \citep{snell2024scaling, chen2025rethinking, dang2025weight}, the relative benefits of instruction tuning versus applying test-time reasoning to base models remain underexplored.


In this work, we address this gap by conducting a controlled empirical comparison between base and instruction-tuned models under comparable inference-time scaling. We evaluate models using the Pass@20 metric across a diverse suite of benchmarks: standard mathematical reasoning tasks (GSM8K \citep{gsm8k_cobbe2021training}, Math-500 \citep{math500_lightman2023lets}), perturbed variants designed to disrupt solution templates (Math-Perturb Hard \citep{math_perturb_huang2025math}), and a domain-specific clinical benchmark, MedCalc \citep{medcalc_khandekar2024medcalc}. Motivated by recent advances in sampling methods, the base models utilize CoT decoding \citep{cot_decoding_wang2024chain} while instruction-tuned models employ repeated stochastic sampling. 


Our contributions are summarized as follows:

\begin{itemize}
    \item \textbf{Base models dominate in zero-shot:} Despite superior few-shot CoT results on GSM8K, instruction-tuned models significantly underperform their base counterparts in zero-shot settings, exhibiting performance drops of 32.7\% (LLaMA3-70B) and 31.2\% (Kimi-K2).
    
    \item \textbf{Poor transferability under distribution shift:} Performance gains from instruction tuning do not reliably transfer to the domain-specific MedCalc benchmark. Base models outperform instruction-tuned counterparts in zero-shot CoT, with observed drops of $6.78$\% (LLaMA3-$70$B) and $6.11$\% (Kimi-K2).
    
    \item \textbf{Sensitivity to perturbation and evaluation noise:} Instruction-tuned models show limited robustness, suffering sharp declines from Math-500 to Math-Perturb Hard (e.g., Kimi-K2 drops from $94.20$\% to $76.34$\%). Additionally, we identify reliability issues in standard rule-based regex evaluation for latex math benchmarks, which lead to scoring discrepancies of up to $10.8$\%.
\end{itemize}

\section{Related Work}

Instruction following in LLMs is primarily established via supervised fine-tuning (SFT) or reinforcement learning (RL). \citet{instruct_gpt_ouyang2022training} demonstrated that aligning base models with human feedback enhances both user intent following and task performance. Subsequent research has scaled this approach through improved training methodologies and the curation of diverse natural \citep{naturalinstructions, supernaturalinstructions, flan_wei2021finetuned} and synthetic datasets \citep{self_instruct_wang2023self, alpaca, han2023medalpaca}. Consequently, instruction-tuned models are often assumed to be strictly superior to their base counterparts. However, with the emergence of massive-scale base models (up to 1 trillion parameters \citep{kimi_k2_team2025kimi}), we revisit this assumption to determine if it holds at scale.

Parallel to instruct-tuning, prompting strategies have significantly improved reasoning capabilities. \citet{cot_prompting_wei2022chain} observed that fewshot CoT prompting with intermediate reasoning steps substantially improves performance on mathematical tasks \citep{gsm8k_cobbe2021training}. Similarly, self-consistency \citep{sefl_consistency_wang2022self} exploits LLM stochasticity to sample multiple trajectories and aggregate results via majority voting. Distinct from repeated sampling, CoT decoding \citep{cot_decoding_wang2024chain} explicitly branches at the first token by selecting top-$K$ candidates, followed by greedy decoding. Notably, this method demonstrates that base models possess substantial latent reasoning ability. Accordingly, we adopt CoT decoding as the default strategy for base models in this work.

The reliance on open-source benchmarks introduces risks of data contamination, where models may memorize evaluation data during training. Recent studies \citep{data_contamination_deng-etal-2024-investigating, dcr_xu-etal-2025-dcr, reasoning_or_memorization_wu2025reasoning} question whether high performance on standard datasets reflects genuine reasoning. To address this, we extend our evaluation beyond standard benchmarks to include Math Perturb \citep{math_perturb_huang2025math}, designed to disrupt solution templates, and the clinical benchmark MedCalc \citep{medcalc_khandekar2024medcalc}. We hypothesize: if instruction-tuned models possess superior reasoning, their advantages must persist under these distribution shifts.

\section{Experiments}
We evaluate the comparative performance of instruction-tuned models and their base counterparts under scalable test-time inference using Pass@$k$ metric ($k=20$).

\textbf{Models:} We examine a suite of 16 models spanning 0.6B to 1T parameters across five prominent families: Qwen \citep{yang2025qwen3}, LLaMA \citep{grattafiori2024llama}, SmolLM \citep{bakouch2025smollm3}, DeepSeek \citep{liu2024deepseek}, and Kimi \citep{kimi_k2_team2025kimi}. We focus exclusively on open-weights models to ensure access to corresponding base checkpoints. Additional implementation details are provided in the \cref{sec:implementation_details_suppl}.

\textbf{Datasets:} To test the hypothesis that instruction tuning gains generalize beyond standard evaluations, we select four diverse datasets. We employ GSM8K \citep{gsm8k_cobbe2021training} and Math-500 \citep{math500_lightman2023lets} as baselines. To assess robustness against structural perturbations, we utilize Math Perturb Hard \citep{math_perturb_huang2025math}, which modifies Math-500 problems specifically to disrupt solution templates and expose reliance on memorized patterns. Finally, we evaluate domain transfer using MedCalc \citep{medcalc_khandekar2024medcalc}, a clinical calculation benchmark comprising over 1,000 manually curated calculation tasks across 55 distinct types.

\section{Results}

\subsection{Comparison on standard benchmarks}
\begin{figure}
    \centering
    \includegraphics[width=0.85\linewidth]{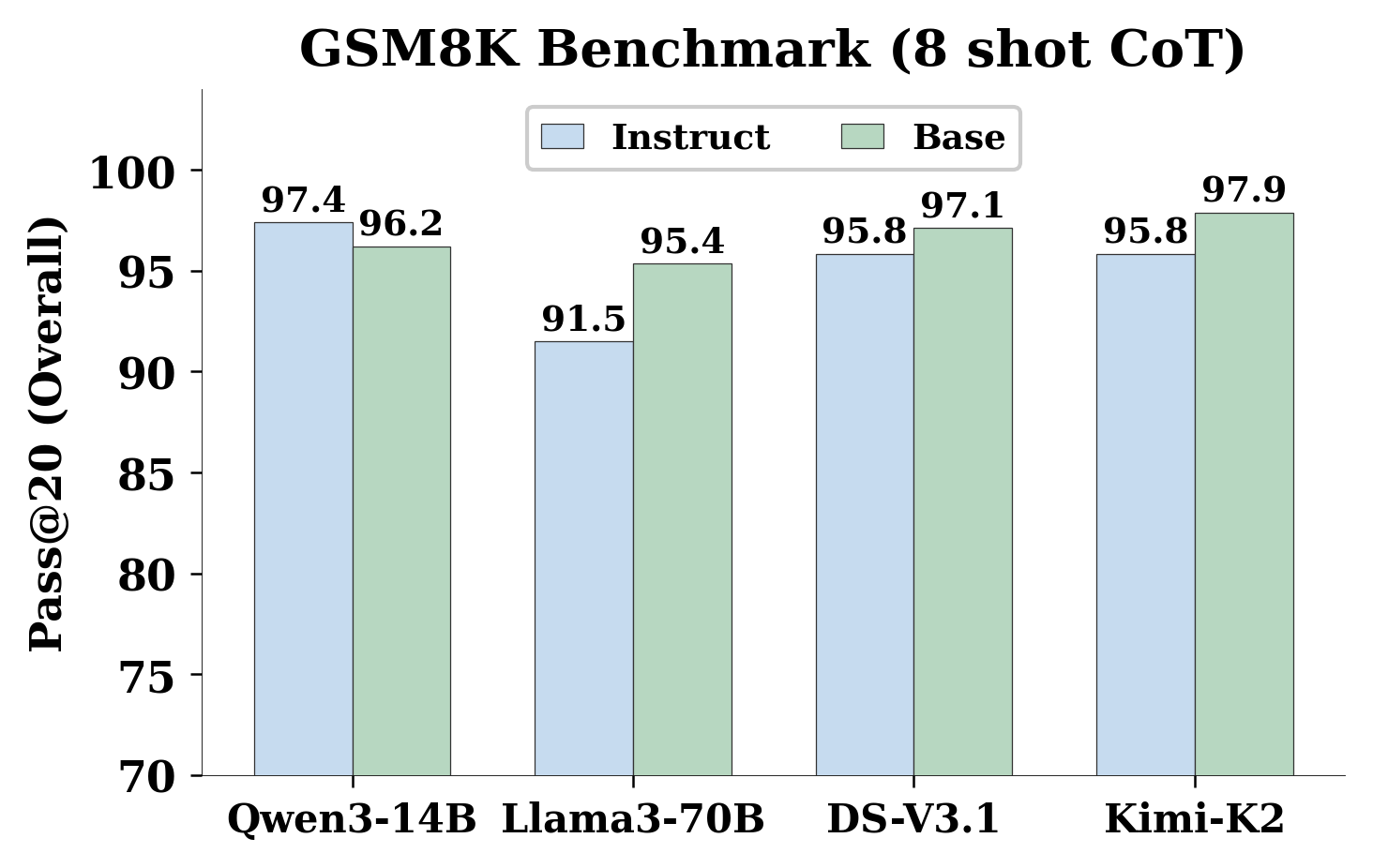}
    \caption{Pass@20 on GSM8K. Instruction tuning yields marginal gains here, with mid-to-large base models often surpassing their instruction-tuned counterparts.}
    \label{fig:gsm8k_8_shot_main}
    \vspace{-1em}
\end{figure}

On GSM8K (8-shot CoT), base and instruction-tuned models perform similarly (Fig.~\ref{fig:gsm8k_8_shot_main}), while most mid/large base models outperform their instruction counterparts, including LLaMA3-70B (95.4\% vs. 91.5\%), DS-V3.1 (97.1\% vs. 95.8\%), and Kimi-K2 (97.9\% vs. 95.8\%). Notably, instruction-tuned models provide marginal benefit on GSM8K (especially in SLMs), with gains diminishing as model size increases (Fig.~\ref{fig:gsm8k_full_appendix}). 
\begin{figure}
    \centering
    \includegraphics[width=0.85\linewidth]{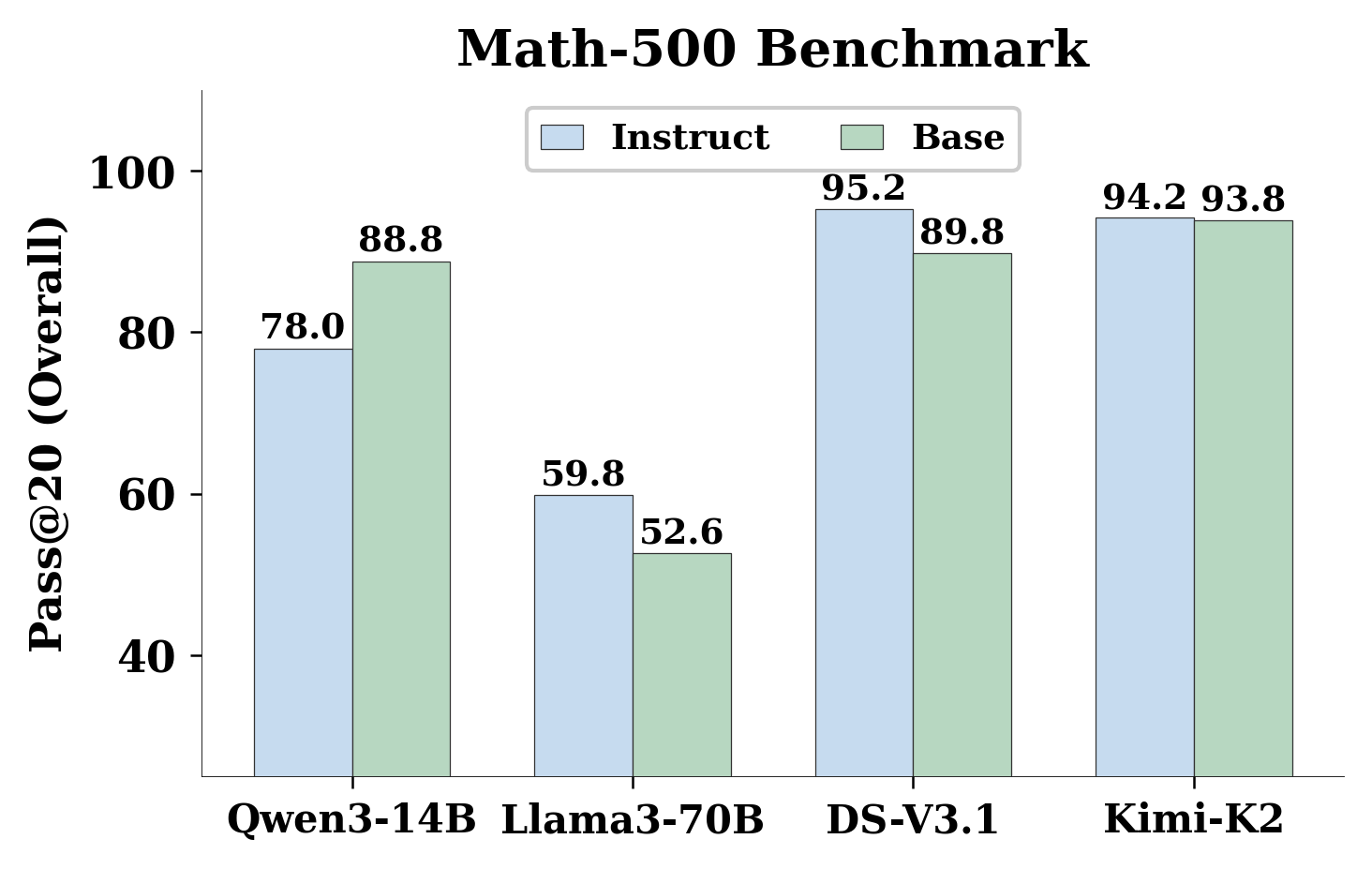}
        \caption{Pass@20 on Math-500. Gap between instruction and base models reducing with parameter scale.
        }
    \label{fig:Math-500_main_fig}
\end{figure}In contrast, instruction tuning has a larger effect on Math-500 for SLMs, but this trend weakens for LLMs (Fig.~\ref{fig:Math-500_main_fig}), where base models often match or exceed instruction-tuned performance. For example, LLaMA3-3B improves from 29.0\% to 67.2\% and SmollM-3B from 72.0\% to 86.2\% (Fig. ~\ref{fig:Math-500_full_appendix}). This suggests that large base models already capture much of the reasoning required for both benchmarks.



\begin{findingbox}
    \textbf{\textit{Finding 1:}} 
    At $>70$B scale, base models perform competitively (Math-500) and superior (GSM8K) relative to instruction-tuned models.
\end{findingbox}

\subsection{Robustness under distribution shift}
\label{sec:comparsion_less_popular_benchmark}

\begin{figure}[h]
    \centering
    \includegraphics[width=0.85\linewidth]{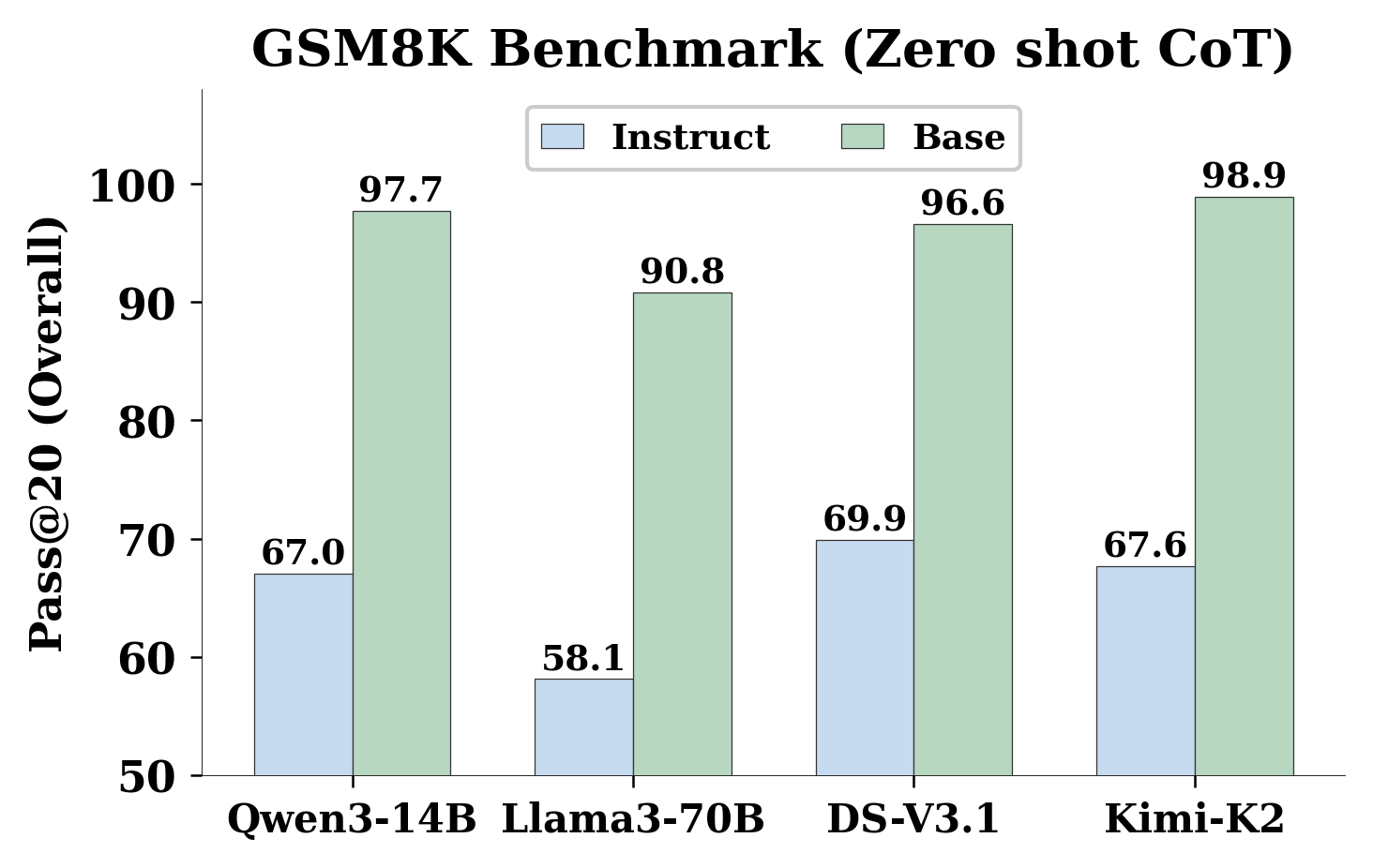}
    
    \caption{Pass@20 on GSM8K (0-shot CoT). Base models significantly outperform instruction-tuned models}
    \label{fig:gsm8k_grader_0_shot_main}
\end{figure}

To assess whether instruction tuning gains persist under domain shift, we evaluate models on MedCalc (Tab~\ref{tab:medcalc_zero_shot_cot_main}). In zero-shot CoT, base models outperform instruction-tuned models across most categories and model sizes, with the largest gaps observed for SLMs (LLaMA3-3B, SmollM-3B).
\begin{table}[h]
\centering
\resizebox{0.999\linewidth}{!}{
\begin{tabular}{lccccccccc}
\toprule
\textbf{Model / Setting} & \textbf{Lab} & \textbf{Risk} & \textbf{Physical} & \textbf{Severity} & \textbf{Diagnosis} & \textbf{Date} & \textbf{Dosage} & \textbf{Overall} \\
\midrule
Qwen3-0.6B   & 15.60 & 39.17 & 40.00 & 25.00 & 60.00 & 28.33 & 10.00 & 30.37 \\
Base      & 30.28 & 65.83 & 36.67 & 48.75 & 93.33 & 5.00 & 10.00 & 42.69 \\
$\Delta$ & \deltacol{-14.68} & \deltacol{-26.66} & \deltacol{3.33} & \deltacol{-23.75} & \deltacol{-33.33} & \deltacol{23.33} & \deltacol{0.00} & \deltacol{-12.32} \\
\midrule
SmollM-3B   & 25.99 & 40.83 & 57.08 & 37.50 & 70.00 & 20.00 & 17.50 & 39.26 \\
Base      & 54.74 & 62.50 & 85.42 & 46.25 & 81.67 & 28.33 & 27.50 & 61.89 \\
$\Delta$ & \deltacol{-28.75} & \deltacol{-21.67} & \deltacol{-28.34} & \deltacol{-8.75} & \deltacol{-11.67} & \deltacol{-8.33} & \deltacol{-10.00} & \deltacol{-22.63} \\
\midrule
Llama3-3B  & 26.30 & 35.83 & 27.50 & 27.50 & 53.33 & 8.33 & 15.00 & 28.94 \\
Base & 54.74 & 62.50 & 86.25 & 46.25 & 81.67 & 28.33 & 27.50 & 62.08 \\
$\Delta$ & \deltacol{-28.44} & \deltacol{-26.67} & \deltacol{-58.75} & \deltacol{-18.75} & \deltacol{-28.34} & \deltacol{-20.00} & \deltacol{-12.50} & \deltacol{-33.14} \\
\midrule
Qwen3-8B  & 57.49 & 50.83 & 76.25 & 28.75 & 55.00 & 60.00 & 25.00 & 56.83 \\
Base  & 34.86 & 64.17 & 75.83 & 37.50 & 88.33 & 10.00 & 15.00 & 52.05 \\
$\Delta$ & \deltacol{22.63} & \deltacol{-13.34} & \deltacol{0.42} & \deltacol{-8.75} & \deltacol{-33.33} & \deltacol{50.00} & \deltacol{10.00} & \deltacol{4.78} \\
\midrule
Qwen3-14B  & 61.47 & 53.75 & 83.33 & 31.25 & 65.00 & 41.67 & 27.50 & 60.17 \\
Base  & 62.69 & 65.00 & 93.33 & 51.25 & 90.00 & 60.00 & 45.00 & 70.11 \\
$\Delta$ & \deltacol{-1.22} & \deltacol{-11.25} & \deltacol{-10.00} & \deltacol{-20.00} & \deltacol{-25.00} & \deltacol{-18.33} & \deltacol{-17.50} & \deltacol{-9.94} \\
\midrule
Llama3-70B  & 38.53 & 35.00 & 69.17 & 23.75 & 66.67 & 35.00 & 35.00 & 44.89 \\
Base  & 43.12 & 49.17 & 76.67 & 40.00 & 83.33 & 5.00 & 32.50 & 51.67 \\
$\Delta$ & \deltacol{-4.59} & \deltacol{-14.17} & \deltacol{-7.50} & \deltacol{-16.25} & \deltacol{-16.66} & \deltacol{30.00} & \deltacol{2.50} & \deltacol{-6.78} \\
\midrule
DS-V3.1   & 69.42 & 55.00 & 81.67 & 58.75 & 66.67 & 41.67 & 55.00 & 65.81 \\
Base      & 55.68 & 56.89 & 83.27 & 40.00 & 68.33 & 71.67 & 60.00 & 62.99 \\
$\Delta$ & \deltacol{13.74} & \deltacol{-1.89} & \deltacol{-1.60} & \deltacol{18.75} & \deltacol{-1.66} & \deltacol{-30.00} & \deltacol{-5.00} & \deltacol{2.82} \\
\midrule
Kimi-K2  & 72.17 & 64.58 & 88.75 & 48.75 & 68.33 & 46.67 & 55.00 & 70.11 \\
Base  & 61.47 & 77.92 & 94.58 & 65.00 & 96.67 & 68.33 & 80.00 & 76.22 \\
$\Delta$ & \deltacol{10.70} & \deltacol{-13.34} & \deltacol{-5.83} & \deltacol{-16.25} & \deltacol{-28.34} & \deltacol{-21.66} & \deltacol{-25.00} & \deltacol{-6.11} \\
\bottomrule
\end{tabular}}
\caption{Pass@20 on MedCalc (zero-shot CoT). Base models outperform instruction-tuned variants across most categories
$\Delta$ shows (Instruct--Base); \textcolor{red}{red} indicates base superiority.}
\label{tab:medcalc_zero_shot_cot_main}
\vspace{-1.5em}
\end{table} 
These results indicate that the benefits of instruction tuning do not always transfer under distribution shift, underscoring the importance of evaluating beyond one domain and reporting base models. 
We further revisit the GSM8K benchmark under zero-shot CoT to assess if the base vs. instruction performance gap persists in the absence of few-shot exemplars, refer Fig.~\ref{fig:gsm8k_grader_0_shot_main}).
Consistent with MedCalc, base models outperform instruction-tuned models, which requires few-shot CoT (n=8; Fig~\ref{fig:gsm8k_8_shot_main}) to reach comparable accuracy. This suggests that instruction tuning may change how reasoning is elicited, increasing sensitivity to CoT exemplars, contrary to common assumption it consistently improves performance over base models. Also while few-shot CoT largely narrows this gap, base models seem to benefit relatively lesser from additional examples, indicating that their reasoning ability is already active in the zero-shot setting. 


\begin{findingbox}
    \textbf{\textit{Finding 2:}} 
    Under distribution shift, base models outperform instruction-tuned models in zero-shot, while instruction-tuned performance relies on few-shot examples.
\end{findingbox}

\subsection{Few-shot effects on base models}
While instruction-tuned models benefit from few-shot exemplars, our results in Table~\ref{tab:zero_shot_vs_few_shot} shows that on both GSM8K and MedCalc, the base models also improves with few-shot CoT. 
\begin{table}[h!]
\centering\resizebox{0.88\linewidth}{!}{
\begin{tabular}{l|cc|cc}
\toprule
\textbf{Base Model} & 
\multicolumn{2}{c|}{GSM8K} & \multicolumn{2}{c}{MedCalc}
\\
& \textbf{Zero-shot} & \textbf{Few-shot} & \textbf{Zero-shot} & \textbf{Few-shot} \\
\midrule
Qwen3-0.6B        & 86.50 & 83.25 & 42.69 & 52.05 \\
Smollm-3B  & 86.50 & 93.25 & 61.89 & 44.79 \\
Llama-3B  & 56.71 & 73.92 & 62.08 & 46.42 \\
Qwen3-8B  & 97.65 & 92.95 & 52.05 & 72.40 \\
Qwen3-14B  & 97.72 & 96.21 & 70.11 & 79.08 \\
Llama3-70B  & 90.82 & 95.37 & 51.67 & 61.60 \\
DS-V3.1  & 96.59 & 97.12 & 62.99 & 78.99 \\
Kimi-K2  & 98.86 & 97.88 & 76.22 & 89.40 \\
\bottomrule
\end{tabular}}
\caption{Pass@20 comparing zero-shot vs. few-shot CoT on base models, showcasing they too (akin to instruction-tuned) improve with few-shot exemplars.}
\label{tab:zero_shot_vs_few_shot}
\end{table}
This aligns with our earlier observations (Sec~\ref{sec:comparsion_less_popular_benchmark}) that larger base models already possess enough knowledge in zero-shot settings. Empirically, on GSM8K benchmark, the base models show higher zero-shot performance, and perhaps, therefore exhibit a weaker reliance on exemplars compared to instruction-tuned models. Also, on the MedCalc benchmark, under 1 shot CoT (Tab~\ref{tab:medcalc_zero_shot_cot_main}), all base models (except Qwen3-0.6B) outperform their instruction counterparts, reinforcing our hypothesis that instruction tuning does not always lead to improved performance.
\begin{findingbox}
    \textbf{\textit{Finding 3:}} Few-shot learning benefits both instruction-tuned and base models; however, base models rely less on few-shot exemplars due to their stronger zero-shot performance
\end{findingbox}
\subsection{Evaluator Sensitivity: Revisit Math-500}
\begin{table}[t!]
\centering
\small
\setlength{\tabcolsep}{2pt}
\begin{tabular}{llcc}
\toprule
\textbf{Model} & \textbf{Variant} & \textbf{Grader} & \textbf{MathVerify} \\
\midrule
\multirow{3}{*}{Qwen3-14B}
 & Instruct & 78.00 & 84.40 \\
 & Base     & 88.80 & 85.20 \\
 & $\Delta$ & \deltacol{-10.80} & \deltacol{-0.80} \\
\midrule
\multirow{3}{*}{LLaMA3-70B}
 & Instruct & 59.80 & 56.80 \\
 & Base     & 52.60 & 53.20 \\
 & $\Delta$ & \deltacol{7.20} & \deltacol{3.60} \\
\midrule
\multirow{3}{*}{DS-V3.1}
 & Instruct & 95.20 & 89.40 \\
 & Base     & 89.80 & 86.00 \\
 & $\Delta$ & \deltacol{5.40} & \deltacol{3.40} \\
\midrule
\multirow{3}{*}{Kimi-K2}
 & Instruct & 94.20 & 88.60 \\
 & Base     & 93.80 & 89.00 \\
 & $\Delta$ & \deltacol{0.40} & \deltacol{-0.40} \\
\bottomrule
\end{tabular}
\caption{Evaluator sensitivity on Math-500 (Pass@20). Performance gaps vary significantly. Due to space constraints, we show manually audited cases demonstrating grader outputs in Sec~\ref{sec:manual_examples_audit} for interested readers.
}
\label{tab:Math-500_grader_mathverify}
\vspace{-2em}
\end{table}
Although Math-500 is a widely cited benchmark, we found evaluating it remains complicated as LLMs need to not only predict correct answer but also format it in valid latex. Therefore, prior work relies on heuristic, regex based answer extraction rules, making evaluations sensitive to formatting rather than solving skill, which is also reflected in substantial discrepancies across evaluators (Tab~\ref{tab:Math-500_grader_mathverify} and Tab~\ref{tab:Math-500_grader_mathverify_appendix}). For example, on Qwen3-14B, the standard grader reports a gap of $10.8\%$, whereas Math Verify reduces this difference to $0.8\%$. Similarly, for Kimi-K2, the reported gap is $0.4\%$, but with opposite directions under the standard grader and MathVerify. Moreover, under the standard Math grader, strong Math-500 performance does not reliably transfer to Math Perturb Hard, with large drops observed for LLMs (e.g., LLaMA3-70B: $59.80\%$ vs. $22.22\%$; full results in Tab\ref{tab:Math-500_vs_math_perturb_appendix}). Together, these results raise concerns about the reliability of Math-500 reported results.

\begin{findingbox}
    \textbf{\textit{Finding 4:}} Gains of instruction-tuned models over base models are found to be highly evaluator dependent on LaTeX based benchmarks and primarily benefit SLMs, with mixed and often marginal effects for LLMs.
\end{findingbox}


\section{Conclusion}
In this work, we revisit the assumption that instruction tuning uniformly improves LLMs performance. With systematic evaluation on standard (GSM8K, Math-500) and domain-shifted and robustness focused benchmarks (MedCalc, Math Perturb), we show that its benefits over base models depend strongly on number of parameters and evaluation setting (zero-shot vs few-shot).
We also observe that instruction tuning provides substantial gains for SLMs, while its impact on LLMs is mixed and often marginal. Our analysis also support concerns about benchmark contamination/memorization, as instruction-tuned models perform noticeably worse on Math Perturb Hard than on Math-500. This highlights the need to interpret instruction tuning gains with caution and to rely on robust,  benchmarks when assessing true reasoning ability in LLMs.

\section*{Limitations}
Our study focuses on comparing base and instruction-tuned variants across a set of widely used math and domain-shifted benchmarks. This scope leaves several limitations.

\textbf{Benchmark coverage} We evaluate a finite set of tasks (e.g., GSM8K, Math-500, and perturbation-based variants). While these benchmarks are standard, they may not fully represent real-world reasoning, and our results may hold true for benchmark-specific artifacts.

\textbf{Evaluator dependence} A central finding is that reported gaps can vary across graders, especially on latex heavy outputs. Although we include multiple evaluators and manual audits, we do not claim to exhaust the space of possible evaluators or formatting behaviors. Further, our manual audits are necessarily limited in size and may miss rare failure modes.

\textbf{Decoding and prompting choices.}
Results depend on generation settings (e.g., max tokens, temperature) and prompt templates (zero-shot/few-shot, CoT formatting). While we keep settings consistent within comparisons, different choices may shift some of our findings.

\textbf{Model and data availability constraints.}
We evaluate a selected set of model families and sizes based on access and compute constraints. Consequently, our conclusions should not be interpreted as universal across all instruction-tuning recipes.

\textbf{No causal claims about instruction tuning} Our analysis is empirical and comparative; it does not isolate which components of instruction tuning (data mixture, loss function, RLHF/DPO variants, formatting conventions) drive the observed effects. Enabling a closer inspection on training-time interventions is beyond the scope of this work. Moreover, instruction-tuned models are commonly assumed to outperform base models due to extensive fine-tuning, here our goal was to examine whether this assumption holds when base models, without instruction tuning, are paired with stronger sampling strategies. We leave the study of applying identical sampling methods to instruction-tuned models to future work, as our focus is not on optimizing performance but on testing the robustness of this widely held assumption.

\section*{Ethical considerations}
This work empirically studies the performance comparisons between base and instruction-tuned models on established and standard benchmarks.

\textbf{Risk of overgeneralization.}
A potential harm is that readers may overinterpret benchmark results as evidence about real-world reliability or safety. We explicitly emphasize evaluator dependence, decoding sensitivity, and benchmark limitations, and we recommend cautious interpretation of apparent gains.

\textbf{Responsible reporting.} Our findings could be misused to selectively choose evaluators that favor a preferred model. To counter this, we encourage reporting results across multiple graders (or robust verification-based evaluation) and documenting decoding settings, extraction rules, and error analyses.

\textbf{Environmental impact.}
Running large-scale evaluations and multiple sampling runs can consume substantial compute. We reduce unnecessary overhead by reusing cached generations where feasible and prioritizing targeted analyses (e.g., perturbation) to answer specific methodological questions.

\bibliography{custom}

\appendix
\newpage

\counterwithin{figure}{section}
\renewcommand{\thefigure}{S\arabic{figure}}

\counterwithin{table}{section}
\renewcommand{\thetable}{S\arabic{table}}

\section{Performance on GSM8K \& Math-500}

\begin{figure*}
    \centering
    \includegraphics[width=0.9\linewidth]{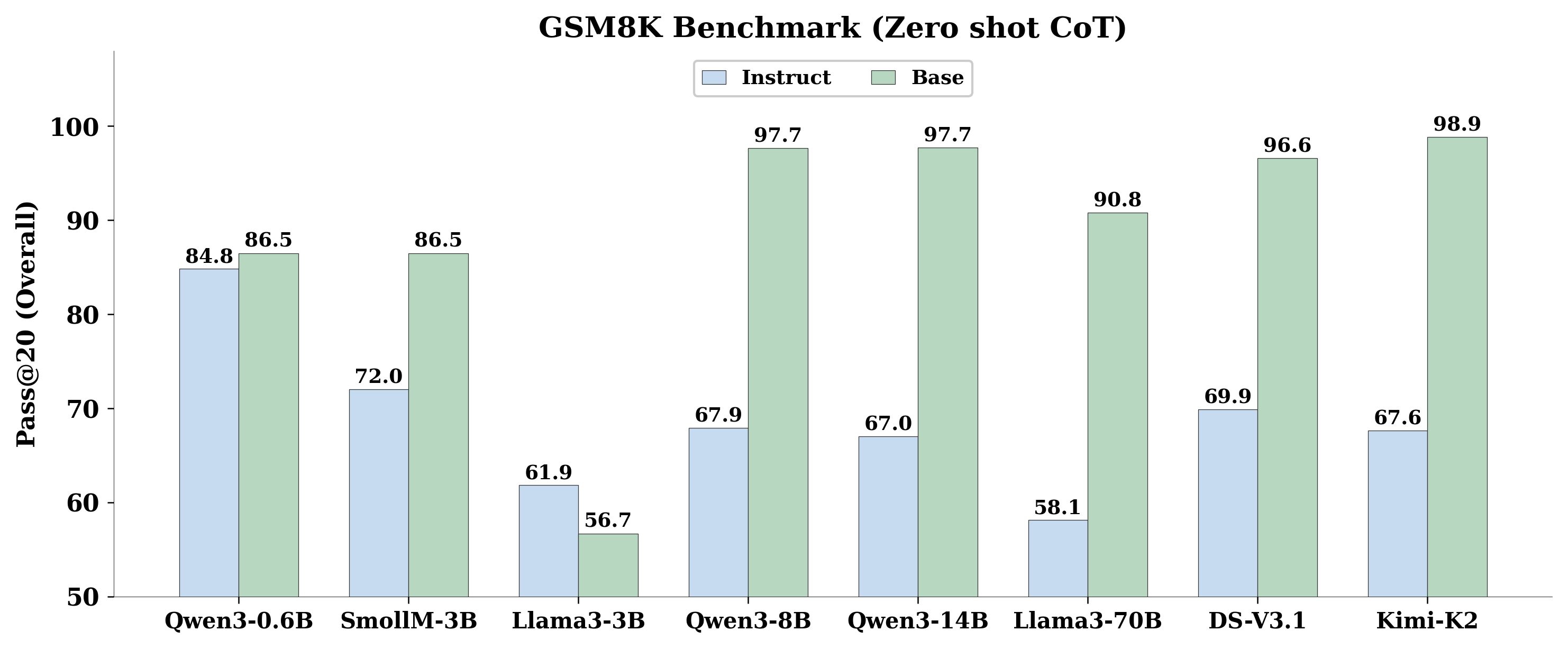}
    \caption{Base models versus Instruct models on GSM8K benchmark under zero-shot CoT setting}
    \label{fig:gsm8k_0_shot_full_appendix}
\end{figure*}

\begin{figure*}
    \centering
    \includegraphics[width=0.9\linewidth]{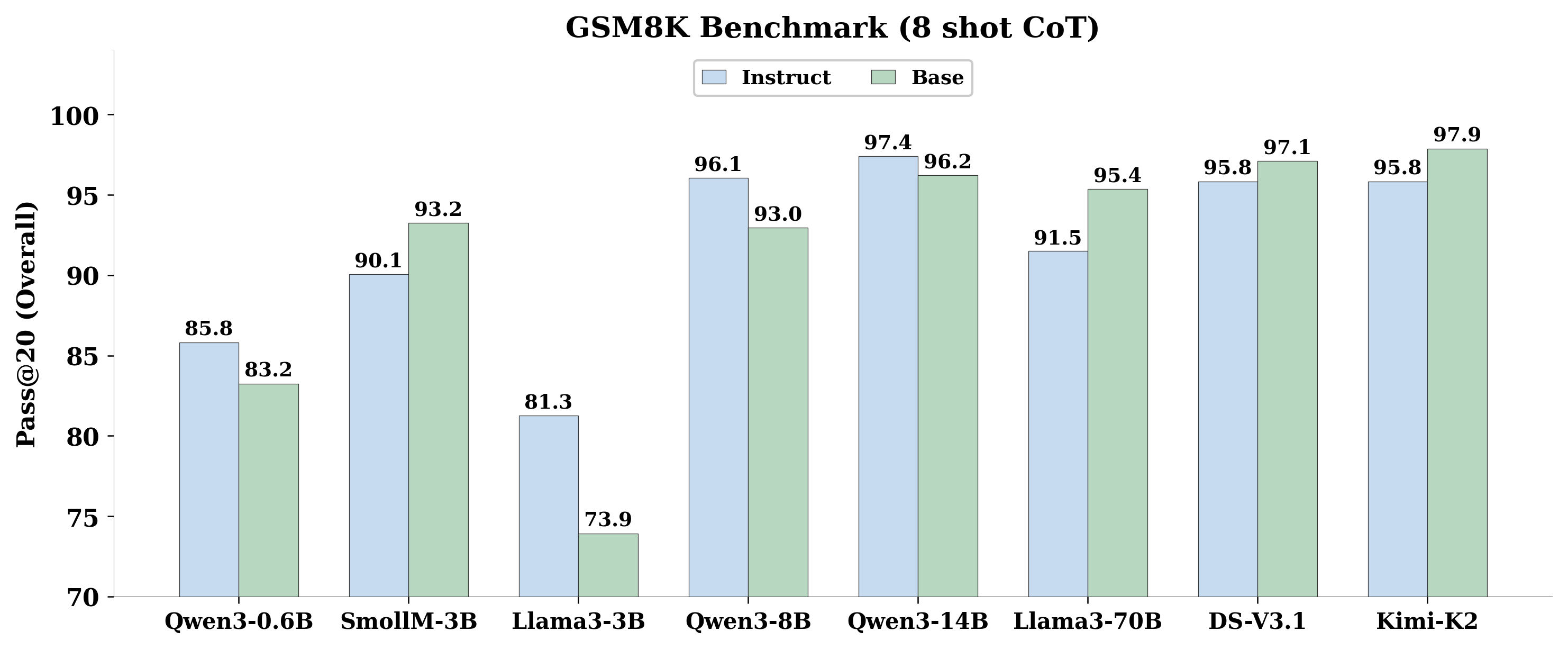}
    \caption{Base models versus Instruct models on GSM8K benchmark under 8-shot CoT setting}
    \label{fig:gsm8k_full_appendix}
\end{figure*}

In this section, we report results for additional models not included in the main paper due to space constraints. Fig\ref{fig:gsm8k_full_appendix}, Tab~\ref{tab:gsm8k_zero_vs_few_appendix} and Fig\ref{fig:gsm8k_0_shot_full_appendix} further support our observation that, under zero-shot evaluation, base models often outperform instruction-tuned models. 

\begin{table}[h!]

\centering\resizebox{0.9\linewidth}{!}{
\begin{tabular}{lcccc}
\toprule
\textbf{Model} &
\multicolumn{2}{c}{Math-500} & \multicolumn{2}{c}{Math Perturb Hard}
\\
 & \textbf{Grader} & \textbf{Math Verify} & \textbf{Grader} & \textbf{Math Verify} \\
\midrule

Qwen3-14B   & 78.00 & 84.40 & 39.78 & 68.10 \\
\midrule
Llama3-70B  & 59.80 & 56.80 & 22.22 & 26.16 \\
\midrule
DS-V3.1     & 95.20 & 89.40 & 77.78 & 77.42 \\
\midrule
Kimi-K2     & 94.20 & 88.60 & 76.34 & 64.87 \\
\bottomrule
\end{tabular}}

\caption{Pass@20 on Math-500 vs. Math Perturb Hard}
\label{tab:Math-500_vs_math_perturb_appendix}
\end{table}

\begin{figure*}
    \centering
    \includegraphics[width=0.9\linewidth]{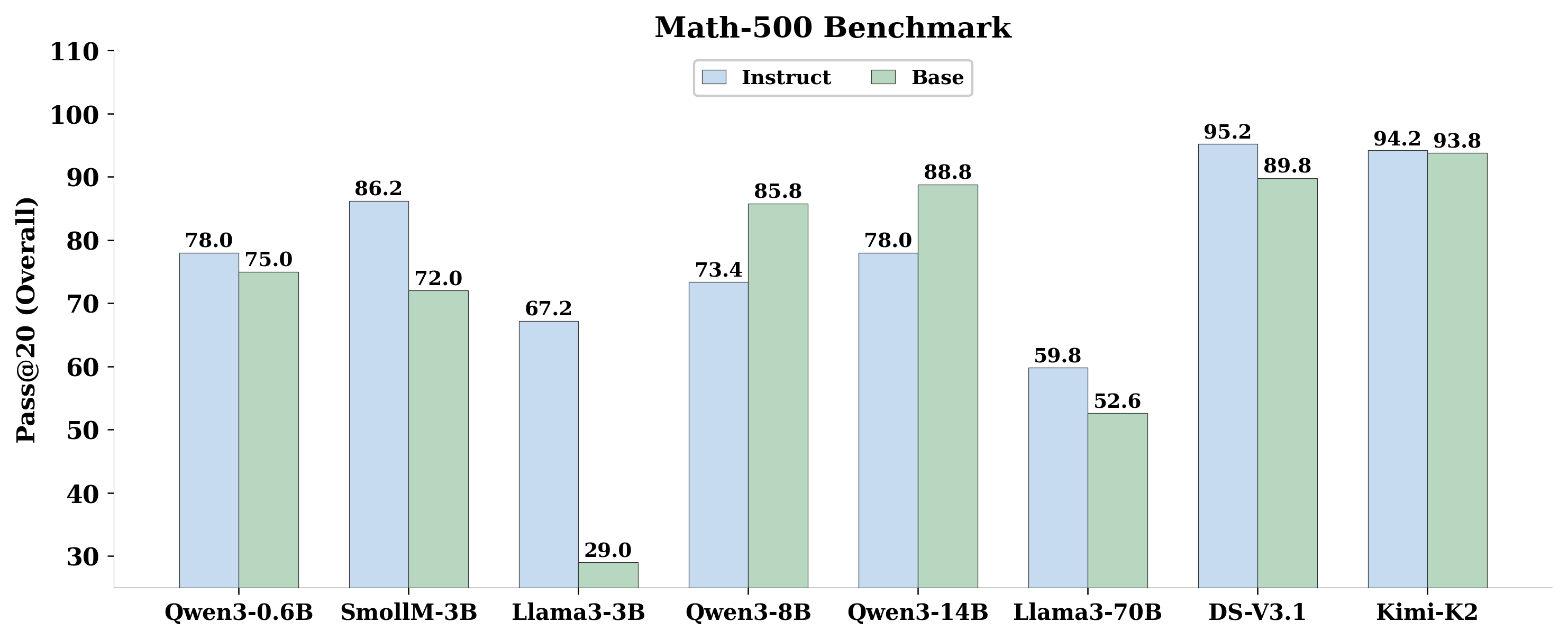}
    \caption{Base models versus Instruct models on Math-500 benchmark. }
    \label{fig:Math-500_full_appendix}
\end{figure*}

For Math-500 (refer Fig~\ref{fig:Math-500_full_appendix}), we find that instruction tuning substantially benefits SLMs. However, these gains are not uniform with increasing model scale and diminish for LLMs such as Kimi-K2: $94.2$\% vs. $93.8$\%. We also observe substantial evaluator dependent variance on the Math-500 benchmark, as shown in Tab~\ref{tab:Math-500_grader_mathverify_appendix}, raising reliability concerns over Math-500 reported results. 

\section{Manually Audited Examples on Math-500}
\label{sec:manual_examples_audit}
\begin{figure*}[t]
    \centering
    \includegraphics[width=0.99\linewidth]{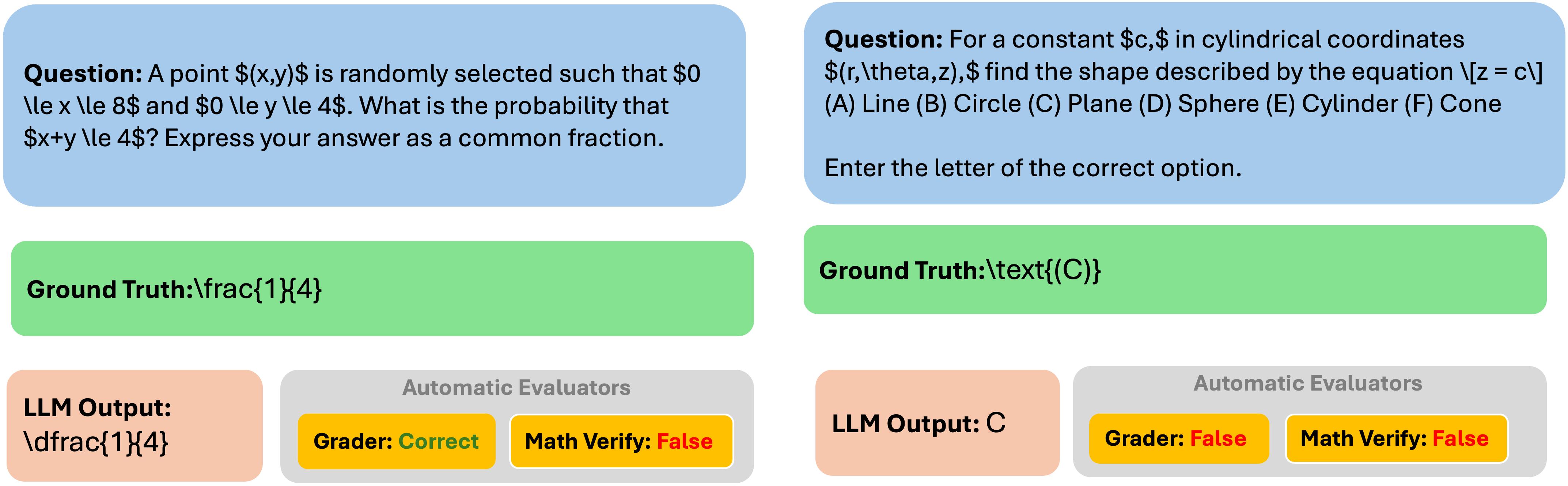}
    \caption{Manually verified examples for Math-500 test set, showing how inconsistent both the grader and math verify are.}
    \label{fig:manual_example_grader_appendix}
\end{figure*}

We manually audit a subset of examples to investigate both the evaluators: Math Verify and the standard grader. In Figure~\ref{fig:manual_example_grader_appendix}, we show the sensitivity of both graders to output variations of LLM. We note that while some of these errors can be mitigated through simple normalization (e.g., replacing \verb|\dfrac| to \verb|\frac|) of LLM outputs, however, rule-based fixes for corner cases will be inherently limited and cannot exhaustively cover all possible formats in which LLM outputs may be expressed.

Refer to Figure~\ref{fig:manual_example_grader_appendix}, in the first case, the 'Math Verify' grader fails to recognize the equivalence between the display style latex fraction \verb|\dfrac{1}{4}| and the ground truth \verb|\frac{1}{4}|, whereas the 'Grader' method successfully handles this syntactic variation. 
In the second case, both evaluation methods fail when the LLM provides the raw answer 'C' instead of the strictly formatted \verb|\text{(C)}|. This highlights a significant limitation in current automated benchmarks: they frequently prioritize exact string or template matching over mathematical or logical equivalence, leading to an inflation of false negative results and an underestimation of the model’s actual reasoning capabilities.

\section{Dataset Statistics}
\label{sec:dataset_statistics}

We evaluate models on four publicly available benchmarks: GSM8K \cite{gsm8k_cobbe2021training}, Math-500 \cite{math500_lightman2023lets}, MedCalc \cite{medcalc_khandekar2024medcalc}, and Math Perturb Hard \cite{math_perturb_huang2025math}. For all datasets, we follow the standard evaluation protocols and use the officially released test splits. No additional training, fine-tuning, or validation-based model selection is performed.


\begin{table}[h!]
\centering
\resizebox{0.99\linewidth}{!}{
\begin{tabular}{lccc}
\toprule
\textbf{Dataset} & \textbf{Domain} & \textbf{Split Used} & \textbf{\# Examples} \\
\midrule
GSM8K & Math (Grade-school) & Test & 1{,}319 \\
Math-500 & Math (Advanced) & Test & 500 \\
MedCalc & Medical Reasoning & Test & 1{,}000 \\
Math Perturb Hard & Math (Robustness) & Test & 279 \\
\bottomrule
\end{tabular}}
\caption{Dataset statistics and evaluation splits}
\label{tab:dataset_stats}
\end{table}

All results reported in the paper are computed on the test sets only. Performance is measured using Pass@\({K}\) with \(K=20\), following prior work. For benchmarks with structured or latex heavy outputs, we adopt robust answer extraction strategies as described in the implementation details (Sec~\ref{sec:implementation_details_suppl}) to ensure consistent evaluation.

\section{Performance on Distribution Shift: MedCalc and Math Perturb}
In addition to the zero-shot results on MedCalc reported in the main paper, we evaluate models under the other two official settings\cite{medcalc_khandekar2024medcalc}: direct-shot and one-shot CoT settings, and report their results in Tab~\ref{tab:medcalc_direct_shot_cot_appendix} and Tab~\ref{tab:medcalc_one_shot_cot_appendix} respectively. Under direct-shot, base models are found in Tab~\ref{tab:medcalc_zero_shot_cot_appendix} to again outperform instruction-tuned models across most categories and almost consistenly in overall performance (except Qwen3-8B). We also observe that instruction-tuned models benefit substantially from zero-shot CoT prompting (i.e., appending \textit{“Let’s think step-by-step”}) compared to direct-shot, whereas base models already achieve strong zero-shot performance. This implies that the knowledge required for solving the MedCalc benchmark is largely present in base models, while instruction-tuned models rely more heavily on prompt structure (direct-shot vs zero-shot) and few-shot exemplars.

Math-500 reports strong performance (refer Tab~\ref{tab:Math-500_grader_mathverify_appendix}) across models, especially for large instruction-tuned models, implying near saturated performance on this benchmark. However, when evaluated on Math-Perturb (Hard set), which is designed to break solution templates while preserving surface level similarity, all models show substantial performance drop (refer Tab~\ref{tab:Math-500_vs_math_perturb_appendix}). For example, Qwen3-14B drops from $78$ to $39.78$, and LLaMA3-70B drops from $59.8$ to $22.2$. These results reinforce our findings that strong performance achieved by instruction-tuned models on standard benchmarks is not robust and reflects limited generalization under distribution shifts. We note that even for top performing instruction-tuned models such as DS-V3.1 ($95.2$ vs $77.8$) and Kimi-K2 ($94.2$ vs $76.3$), high Math-500 accuracy does not reliably transfer to Math-Perturb.

\section{Implementation Details}
\label{sec:implementation_details_suppl}
In this section, we describe the implementation details. For base models, we employ CoT decoding \citep{cot_decoding_wang2024chain} and generate $K(=20)$ samples to compute Pass@$K$ metric. Given the two limitations for base models: (i) they are not trained explicitly to follow user instructions and adhere to output formatting, and (ii) recent works in literature has shown that evaluation outcomes can be sensitive to how we extract answers \citep{jo2025finding, xverify_chen2025xverify, yu2024xfinder}. 

To address this, we employ a lightweight auxiliary LLM to extract the final answer from each generated output. We deliberately choose a SLM (Qwen3-0.6B was used in this study) to keep computational overhead low. This auxiliary LLM acts solely as an information extraction tool and is tasked only with extracting the answer (refer below Auxiliarly LLM prompt). Notably, we do not provide the original question to the auxiliary LLM to avoid biasing the base model’s response. 

Finally, for all experiments, we deployed the models locally using the vLLM \citep{vllm_kwon2023efficient} framework on a single compute node equipped with 8x NVIDIA H200 GPUs.

\begin{tcolorbox}[breakable, title=Auxiliarly LLM Prompt]
{\small
You are an information extraction system. \\

Your task is to read the following text and extract **only the first final answer found**.\\

Rules:\\
1. Output only the first answer in JSON format:\\
{"answer": "<the answer>"}\\

2. If no answer is present, output:\\
{"answer": "none"}\\

3. Ignore everything else in the text, including other questions, answers, or examples.\\

4. Do not include explanations, quotes, or extra text. \\

Output must be valid JSON.
}
\end{tcolorbox}

For instruction-tuned models, we use repeated sampling with $K=20$ generations under standard sampling settings, including a maximum token length of $8192$ and a temperature of $0.05$ (vs $0.0$ used by \citep{yang2025towards}) to introduce stochasticity. Unlike base models, we explicitly prompt instruction-tuned models to produce outputs in a structured JSON format. To handle occasional malformed JSON outputs (such as missing closing braces $\}$ or extra opening/closing brace), we apply a JSON repair library\footnote{\url{https://github.com/mangiucugna/json_repair}}. In practice, we found this step helpful for reliably recovering valid outputs in non-trivial cases.

\begin{tcolorbox}[breakable, title=Instruction-tuned models Prompt]
{\small
You are a helpful assistant for solving math problems.\\

Please think step-by-step to solve the question and then generate the required score.\\

Your output should only contain a JSON dict formatted as \\
\{\\
  ``step\_by\_step\_thinking'':\\str(your\_thinking\_procress\_to\_solve\_the\_question),\\
  ``answer'':\\
str(short\_and\_direct\_answer\_of\_the\_question)\\
\}
}
\end{tcolorbox}
\clearpage
\section{Figures and Tabular Results on GSM8K, Math-500 and Math-Perturb Benchmark}

\begin{table}[h!]
\centering\resizebox{0.99\linewidth}{!}{
\begin{tabular}{lcc}
\toprule
\textbf{Model} & 
\multicolumn{2}{c}{GSM8K}
\\
& \textbf{zero-shot CoT} & \textbf{8-shot CoT} \\
\midrule
Qwen3-0.6B  & 84.84 & 85.82 \\
Base        & 86.50 & 83.25 \\
$\Delta$ & \deltacol{-1.66} & \deltacol{2.57} \\
\midrule
Smollm-3B  & 72.02 & 90.07 \\
Base        & 86.50 & 93.25 \\
$\Delta$ & \deltacol{-14.48} & \deltacol{-3.18} \\
\midrule
Llama-3B  & 61.86 & 81.27 \\
Base        & 56.71 & 73.92 \\
$\Delta$ & \deltacol{5.15} & \deltacol{7.35} \\
\midrule
Qwen3-8B  & 67.93 & 96.06 \\
Base        & 97.65 & 92.95 \\
$\Delta$ & \deltacol{-29.72} & \deltacol{3.11} \\
\midrule
Qwen3-14B  & 67.02 & 97.42 \\
Base        & 97.72 & 96.21 \\
$\Delta$ & \deltacol{-30.70} & \deltacol{1.21} \\
\midrule
Llama3-70B  & 58.15 & 91.51 \\
Base        & 90.82 & 95.37 \\
$\Delta$ & \deltacol{-32.67} & \deltacol{-3.86} \\
\midrule
DS-V3.1  & 69.90 & 95.83 \\
Base        & 96.59 & 97.12 \\
$\Delta$ & \deltacol{-26.69} & \deltacol{-1.29} \\
\midrule
Kimi-K2  & 67.63 & 95.83 \\
Base        & 98.86 & 97.88 \\
$\Delta$ & \deltacol{-31.23} & \deltacol{-2.05} \\
\bottomrule
\end{tabular}}
\caption{Pass@20 results for benchmark GSM8K: Zero-shot and 8-shot CoT}
\label{tab:gsm8k_zero_vs_few_appendix}
\end{table}

\begin{table}[h!]

\centering\resizebox{0.9\linewidth}{!}{
\begin{tabular}{lcc}
\toprule
\textbf{Model} &
\multicolumn{2}{c}{Evaluation Method}
\\
 & \textbf{Grader} & \textbf{Math Verify} \\
\midrule
Qwen3-0.6B  & 78.00 & 79.00  \\
Base        & 75.00 & 73.80 \\
$\Delta$ & \deltacol{3.00} & \deltacol{5.20}  \\
\midrule
Smollm-3B   & 86.20 & 81.80 \\
Base        & 72.00 & 70.20 \\
$\Delta$ & \deltacol{14.20} & \deltacol{11.60} \\
\midrule
Llama-3B    & 67.20 & 64.20 \\
Base        & 29.00 & 33.60 \\
$\Delta$ & \deltacol{38.20} & \deltacol{30.60} \\
\midrule
Qwen3-8B    & 73.40 & 83.00 \\
Base        & 85.80 & 83.00\\
$\Delta$ & \deltacol{-12.40} & \deltacol{0.00}  \\
\midrule
Qwen3-14B   & 78.00 & 84.40 \\
Base        & 88.80 & 85.20 \\
$\Delta$ & \deltacol{-10.80} & \deltacol{-0.80} \\
\midrule
Llama3-70B  & 59.80 & 56.80 \\
Base        & 52.60 & 53.20 \\
$\Delta$ & \deltacol{7.20} & \deltacol{3.60} \\
\midrule
DS-V3.1     & 95.20 & 89.40 \\
Base        & 89.80 & 86.00 \\
$\Delta$ & \deltacol{5.40} & \deltacol{3.40} \\
\midrule
Kimi-K2     & 94.20 & 88.60 \\
Base        & 93.80 & 89.00 \\
$\Delta$ & \deltacol{0.40} & \deltacol{-0.40} \\
\bottomrule
\end{tabular}}

\caption{Pass@20 on Math-500 using different evaluators: Math grader, Math Verify.}
\label{tab:Math-500_grader_mathverify_appendix}
\end{table}


\begin{table*}[h]
\centering
\resizebox{0.7\linewidth}{!}{
\begin{tabular}{l|cccccccc|c}
\toprule
\textbf{Model / Setting} & \textbf{Lab} & \textbf{Risk} & \textbf{Physical} & \textbf{Severity} & \textbf{Diagnosis} & \textbf{Date} & \textbf{Dosage} & \textbf{Overall} \\
\midrule
Qwen3-0.6B   & 11.93 & 37.08 & 30.00 & 18.75 & 73.33 & 16.67 & 7.50 & 25.98 \\
Base      & 30.28 & 65.83 & 36.67 & 48.75 & 93.33 & 5.00 & 10.00 & 42.69 \\
$\Delta$ & \deltacol{-18.35} & \deltacol{-28.75} & \deltacol{-6.67} & \deltacol{-30.00} & \deltacol{-20.00} & \deltacol{11.67} & \deltacol{-2.50} & \deltacol{-16.71} \\
\midrule
SmollM-3B   & 22.02 & 45.83 & 48.75 & 37.50 & 73.33 & 6.67 & 25.00 & 36.96 \\
Base      & 54.74 & 62.50 & 85.42 & 46.25 & 81.67 & 28.33 & 27.50 & 61.89 \\
$\Delta$ & \deltacol{-32.72} & \deltacol{-16.67} & \deltacol{-36.67} & \deltacol{-8.75} & \deltacol{-8.34} & \deltacol{-21.66} & \deltacol{-2.50} & \deltacol{-24.93} \\
\midrule
Llama3-3B  & 18.65 & 13.75 & 19.58 & 17.50 & 13.33 & 3.33 & 10.00 & 16.14 \\
Base      & 54.74 & 62.50 & 86.25 & 46.25 & 81.67 & 28.33 & 27.50 & 62.08 \\
$\Delta$ & \deltacol{-36.09} & \deltacol{-48.75} & \deltacol{-66.67} & \deltacol{-28.75} & \deltacol{-68.34} & \deltacol{-25.00} & \deltacol{-17.50} & \deltacol{-45.94} \\
\midrule
Qwen3-8B  & 56.57 & 46.67 & 77.92 & 25.00 & 51.67 & 61.67 & 22.50 & 55.49 \\
Base  & 34.86 & 64.17 & 75.83 & 37.50 & 88.33 & 10.00 & 15.00 & 52.05 \\
$\Delta$ & \deltacol{21.71} & \deltacol{-17.50} & \deltacol{2.09} & \deltacol{-12.50} & \deltacol{-36.66} & \deltacol{51.67} & \deltacol{7.50} & \deltacol{3.44} \\
\midrule
Qwen3-14B  & 57.80 & 53.33 & 83.33 & 33.75 & 76.67 & 41.67 & 37.50 & 60.17 \\
Base  & 62.69 & 65.00 & 93.33 & 51.25 & 90.00 & 60.00 & 45.00 & 70.11 \\
$\Delta$ & \deltacol{-4.89} & \deltacol{-11.67} & \deltacol{-10.00} & \deltacol{-17.50} & \deltacol{-13.33} & \deltacol{-18.33} & \deltacol{-7.50} & \deltacol{-9.94} \\
\midrule
Llama3-70B  & 19.27 & 16.67 & 33.33 & 13.75 & 36.67 & 13.33 & 10.00 & 21.78 \\
Base  & 43.12 & 49.17 & 76.67 & 40.00 & 83.33 & 5.00 & 32.50 & 51.67 \\
$\Delta$ & \deltacol{-23.85} & \deltacol{-32.50} & \deltacol{-43.34} & \deltacol{-26.25} & \deltacol{-46.66} & \deltacol{8.33} & \deltacol{-22.50} & \deltacol{-29.89} \\
\midrule
DS-V3.1   & 38.53 & 25.83 & 60.42 & 25.00 & 40.00 & 56.67 & 42.50 & 40.88 \\
Base      & 55.68 & 56.89 & 83.27 & 40.00 & 68.33 & 71.67 & 60.00 & 62.99 \\
$\Delta$  & \deltacol{-17.15} & \deltacol{-31.06} & \deltacol{-22.85} & \deltacol{-15.0} & \deltacol{-28.33} & \deltacol{-15.0} & \deltacol{-17.50} & \deltacol{-22.11} \\
\midrule
Kimi-K2  & 34.25 & 25.42 & 61.67 & 18.75 & 41.67 & 26.67 & 52.50 & 38.01 \\
Base  & 61.47 & 77.92 & 94.58 & 65.00 & 96.67 & 68.33 & 80.00 & 76.22 \\
$\Delta$ & \deltacol{-27.22} & \deltacol{-52.50} & \deltacol{-32.91} & \deltacol{-46.25} & \deltacol{-55.00} & \deltacol{-41.66} & \deltacol{-27.50} & \deltacol{-38.21} \\
\bottomrule
\end{tabular}}
\caption{Pass@20 on MedCalc (direct-shot CoT). $\Delta$ shows the improvement of instruction-tuned over base models; \textcolor{red}{red} indicates higher base-model performance, while \textcolor{green!70!black}{green} indicates higher instruction-tuned performance. The base models outperform instruction-tuned variants across most categories.}
\label{tab:medcalc_direct_shot_cot_appendix}
\end{table*}

\begin{table*}
\centering
\resizebox{0.7\linewidth}{!}{
\begin{tabular}{lccccccccc}
\toprule
\textbf{Model / Setting} & \textbf{Lab} & \textbf{Risk} & \textbf{Physical} & \textbf{Severity} & \textbf{Diagnosis} & \textbf{Date} & \textbf{Dosage} & \textbf{Overall} \\
\midrule
Qwen3-0.6B   & 15.60 & 39.17 & 40.00 & 25.00 & 60.00 & 28.33 & 10.00 & 30.37 \\
Base      & 30.28 & 65.83 & 36.67 & 48.75 & 93.33 & 5.00 & 10.00 & 42.69 \\
$\Delta$ & \deltacol{-14.68} & \deltacol{-26.66} & \deltacol{3.33} & \deltacol{-23.75} & \deltacol{-33.33} & \deltacol{23.33} & \deltacol{0.00} & \deltacol{-12.32} \\
\midrule
SmollM-3B   & 25.99 & 40.83 & 57.08 & 37.50 & 70.00 & 20.00 & 17.50 & 39.26 \\
Base      & 54.74 & 62.50 & 85.42 & 46.25 & 81.67 & 28.33 & 27.50 & 61.89 \\
$\Delta$ & \deltacol{-28.75} & \deltacol{-21.67} & \deltacol{-28.34} & \deltacol{-8.75} & \deltacol{-11.67} & \deltacol{-8.33} & \deltacol{-10.00} & \deltacol{-22.63} \\
\midrule
Llama3-3B  & 26.30 & 35.83 & 27.50 & 27.50 & 53.33 & 8.33 & 15.00 & 28.94 \\
Base & 54.74 & 62.50 & 86.25 & 46.25 & 81.67 & 28.33 & 27.50 & 62.08 \\
$\Delta$ & \deltacol{-28.44} & \deltacol{-26.67} & \deltacol{-58.75} & \deltacol{-18.75} & \deltacol{-28.34} & \deltacol{-20.00} & \deltacol{-12.50} & \deltacol{-33.14} \\
\midrule
Qwen3-8B  & 57.49 & 50.83 & 76.25 & 28.75 & 55.00 & 60.00 & 25.00 & 56.83 \\
Base  & 34.86 & 64.17 & 75.83 & 37.50 & 88.33 & 10.00 & 15.00 & 52.05 \\
$\Delta$ & \deltacol{22.63} & \deltacol{-13.34} & \deltacol{0.42} & \deltacol{-8.75} & \deltacol{-33.33} & \deltacol{50.00} & \deltacol{10.00} & \deltacol{4.78} \\
\midrule
Qwen3-14B  & 61.47 & 53.75 & 83.33 & 31.25 & 65.00 & 41.67 & 27.50 & 60.17 \\
Base  & 62.69 & 65.00 & 93.33 & 51.25 & 90.00 & 60.00 & 45.00 & 70.11 \\
$\Delta$ & \deltacol{-1.22} & \deltacol{-11.25} & \deltacol{-10.00} & \deltacol{-20.00} & \deltacol{-25.00} & \deltacol{-18.33} & \deltacol{-17.50} & \deltacol{-9.94} \\
\midrule
Llama3-70B  & 38.53 & 35.00 & 69.17 & 23.75 & 66.67 & 35.00 & 35.00 & 44.89 \\
Base  & 43.12 & 49.17 & 76.67 & 40.00 & 83.33 & 5.00 & 32.50 & 51.67 \\
$\Delta$ & \deltacol{-4.59} & \deltacol{-14.17} & \deltacol{-7.50} & \deltacol{-16.25} & \deltacol{-16.66} & \deltacol{30.00} & \deltacol{2.50} & \deltacol{-6.78} \\
\midrule
DS-V3.1   & 69.42 & 55.00 & 81.67 & 58.75 & 66.67 & 41.67 & 55.00 & 65.81 \\
Base      & 55.68 & 56.89 & 83.27 & 40.00 & 68.33 & 71.67 & 60.00 & 62.99 \\
$\Delta$ & \deltacol{13.74} & \deltacol{-1.89} & \deltacol{-1.60} & \deltacol{18.75} & \deltacol{-1.66} & \deltacol{-30.00} & \deltacol{-5.00} & \deltacol{2.82} \\
\midrule
Kimi-K2  & 72.17 & 64.58 & 88.75 & 48.75 & 68.33 & 46.67 & 55.00 & 70.11 \\
Base  & 61.47 & 77.92 & 94.58 & 65.00 & 96.67 & 68.33 & 80.00 & 76.22 \\
$\Delta$ & \deltacol{10.70} & \deltacol{-13.34} & \deltacol{-5.83} & \deltacol{-16.25} & \deltacol{-28.34} & \deltacol{-21.66} & \deltacol{-25.00} & \deltacol{-6.11} \\
\bottomrule
\end{tabular}}
\caption{Pass@20 on MedCalc (zero-shot CoT). $\Delta$ shows the improvement of instruction-tuned over base models; \textcolor{red}{red} indicates higher base-model performance, while \textcolor{green!70!black}{green} indicates higher instruction-tuned performance. The base models outperform instruction-tuned variants across most categories.}
\label{tab:medcalc_zero_shot_cot_appendix}
\end{table*}

\begin{table*}[ht]
\centering
\resizebox{0.7\linewidth}{!}{
\begin{tabular}{lccccccccc}
\toprule
\textbf{Model / Setting} & \textbf{Lab} & \textbf{Risk} & \textbf{Physical} & \textbf{Severity} & \textbf{Diagnosis} & \textbf{Date} & \textbf{Dosage} & \textbf{Overall} \\
\midrule
Qwen3-0.6B   & 50.46 & 47.92 & 88.33 & 27.50 & 80.00 & 53.33 & 7.50 & 57.02 \\
Base & 34.86 & 64.17 & 75.83 & 37.50 & 88.33 & 10.00 & 15.00 & 52.05 \\
$\Delta$ & \deltacol{15.60} & \deltacol{-16.25} & \deltacol{12.50} & \deltacol{-10.00} & \deltacol{-8.33} & \deltacol{43.33} & \deltacol{-7.50} & \deltacol{4.97} \\
\midrule
SmollM-3B   & 52.60 & 32.92 & 54.58 & 26.25 & 56.67 & 30.00 & 7.50 & 43.74 \\
Base      & 30.28 & 57.08 & 52.92 & 46.25 & 83.33 & 20.00 & 17.50 & 44.79 \\
$\Delta$ & \deltacol{22.32} & \deltacol{-24.16} & \deltacol{1.66} & \deltacol{-20.00} & \deltacol{-26.66} & \deltacol{10.00} & \deltacol{-10.00} & \deltacol{-1.05} \\
\midrule
Llama3-3B  & 46.48 & 22.92 & 37.08 & 25.00 & 46.67 & 20.00 & 15.00 & 34.57 \\
Base & 42.20 & 51.67 & 52.92 & 37.50 & 81.67 & 20.00 & 15.00 & 46.42 \\
$\Delta$ & \deltacol{4.28} & \deltacol{-28.75} & \deltacol{-15.84} & \deltacol{-12.50} & \deltacol{-35.00} & \deltacol{0.00} & \deltacol{0.00} & \deltacol{-11.85} \\
\midrule
Qwen3-8B  & 70.34 & 52.92 & 97.92 & 27.50 & 68.33 & 76.67 & 22.50 & 67.81 \\
Base  & 76.15 & 67.08 & 90.83 & 52.50 & 91.67 & 40.00 & 22.50 & 72.40 \\
$\Delta$ & \deltacol{-5.81} & \deltacol{-14.16} & \deltacol{7.09} & \deltacol{-25.00} & \deltacol{-23.34} & \deltacol{36.67} & \deltacol{0.00} & \deltacol{-4.59} \\
\midrule
Qwen3-14B  & 83.79 & 60.42 & 97.92 & 45.00 & 76.67 & 75.00 & 42.50 & 76.22 \\
Base  & 85.32 & 68.33 & 96.25 & 61.25 & 90.00 & 60.00 & 37.50 & 79.08 \\
$\Delta$ & \deltacol{-1.53} & \deltacol{-7.91} & \deltacol{1.67} & \deltacol{-16.25} & \deltacol{-13.33} & \deltacol{15.00} & \deltacol{5.00} & \deltacol{-2.86} \\
\midrule
Llama3-70B  & 47.71 & 27.08 & 61.25 & 25.00 & 48.33 & 31.67 & 22.50 & 42.50 \\
Base  & 54.74 & 66.25 & 73.33 & 52.50 & 90.00 & 41.67 & 25.00 & 61.60 \\
$\Delta$ & \deltacol{-7.03} & \deltacol{-39.17} & \deltacol{-12.08} & \deltacol{-27.50} & \deltacol{-41.67} & \deltacol{-10.00} & \deltacol{-2.50} & \deltacol{-19.10} \\
\midrule
DS-V3.1   & 82.57 & 60.83 & 95.42 & 50.00 & 73.33 & 51.67 & 57.50 & 74.79 \\
Base      & 82.87 & 71.25 & 96.25 & 62.50 & 91.67 & 60.00 & 32.50 & 78.99 \\
$\Delta$ & \deltacol{-0.30} & \deltacol{-10.42} & \deltacol{-0.83} & \deltacol{-12.50} & \deltacol{-18.34} & \deltacol{-8.33} & \deltacol{25.00} & \deltacol{-4.20} \\
\midrule
Kimi-K2  & 80.12 & 59.58 & 97.08 & 51.25 & 65.00 & 73.33 & 52.50 & 74.79 \\
Base  & 88.69 & 87.92 & 98.75 & 85.00 & 100.00 & 75.00 & 62.50 & 89.40 \\
$\Delta$ & \deltacol{-8.57} & \deltacol{-28.34} & \deltacol{-1.67} & \deltacol{-33.75} & \deltacol{-35.00} & \deltacol{-1.67} & \deltacol{-10.00} & \deltacol{-14.61} \\
\bottomrule
\end{tabular}}
\caption{Pass@20 on MedCalc (one-shot CoT). $\Delta$ shows the improvement of instruction-tuned over base models; \textcolor{red}{red} indicates higher base-model performance, while \textcolor{green!70!black}{green} indicates higher instruction-tuned performance. The base models outperform instruction-tuned variants across most categories.}
\label{tab:medcalc_one_shot_cot_appendix}
\end{table*}

\newpage

\end{document}